# Modelling prosodic structure using Artificial Neural Networks

Jean-Philippe Bernardy[1] Charalambos Themistocleous[1]
[1]CLASP, University of Gothenburg, Sweden

## Abstract

The ability to accurately perceive whether a speaker is asking a question or is making a statement is crucial for any successful interaction. However, learning and classifying tonal patterns has been a challenging task for automatic speech recognition and for models of tonal representation, as tonal contours are characterized by significant variation. This paper provides a classification model of Cypriot Greek questions and statements. We evaluate two state-of-the-art network architectures: a Long Short-Term Memory (LSTM) network and a convolutional network (ConvNet). The ConvNet outperforms the LSTM in the classification task and exhibited an excellent performance with 95% classification accuracy.

Key words: Cypriot Greek, statements, questions, convolutional networks, LSTMs.

## Introduction

The aim of an intonational model is to identify and represent the tonal structure of an utterance for the purposes of classification or generation of tonal contours. Each sentence is characterized by distinct $F0$ patterns that are perceived as different melodies, such as questions, statements, commands, and requests. The purpose of this study is to provide a classification model of Cypriot Greek statements and questions, using information from their fundamental frequency (F0) contours (for a description of Cypriot Greek intonation see Themistocleous, 2011, 2016). To this purpose two different neural network architectures were tested: a Long short-term memory (LSTM) neural network and a convolutional neural network (ConvNet).

**Properties of Tonal Contours**

In statements with broad focus, each phonological word associates with rising and falling tunes; the whole melody starts usually at a higher frequency and declines gradually to a lower frequency (see Figure 1 upper right panel). By contrast, Wh-questions are characterized either by sustained tune at a high frequency (see Figure 1 upper left panel) or by high-low tune at the Wh-word followed by a fall and an optional rise at the end of the utterance (Figure 1 lower panel). Notably, the automatic classification of intonation is challenging as tonal contours can differ depending on linguistic and non-linguistic factors including the segmental structure of utterances, the distribution of lexical stresses, the speech rate, and the physiology of speakers (gender, age, etc.). These factors affect the overall structure of tonal patterns, such as the frequency range and the length of the utterance.

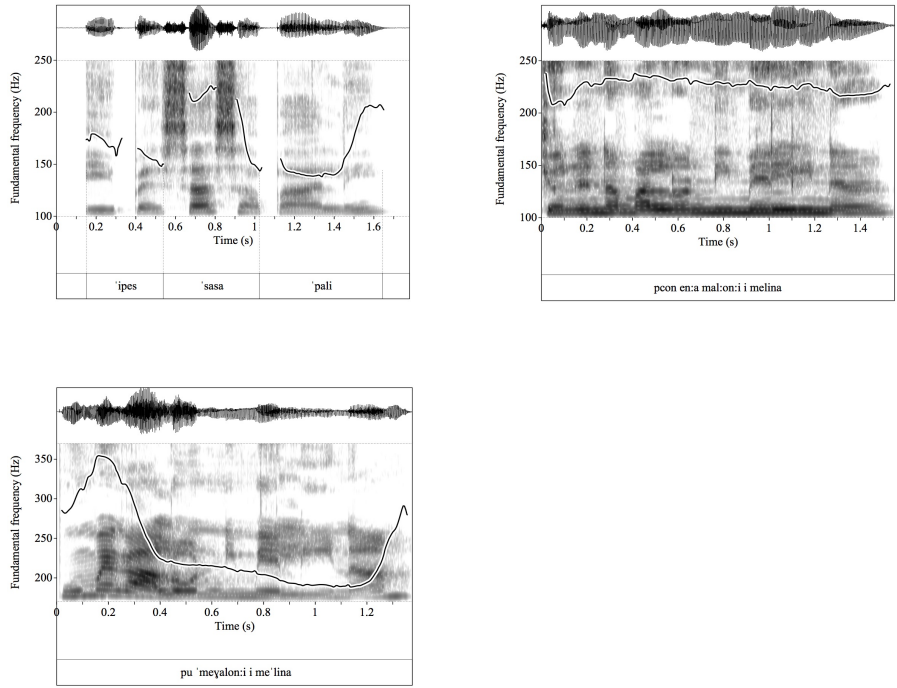

Figure 1. Examples of waveform and spectrogram with superimposed F0 contours of the statement /ˈipes ˈsasa ˈpali/ *you said sasa again* and the wh-questions /pcon eˈnːa maˈlːonːi/ *whom is (s)he going to scold* and /pu meɣa ˈloni i meˈlina/ *where does Melina grow up* uttered by female speakers.

## Artificial Network Architectures

LSTMs are recurrent neural architectures (RNN) (Hochreiter and Schmidhuber, 1997), which proved to be extremely powerful and dynamical systems in recognizing, processing, and predicting tasks such as time series, which are characterized by time lags of unknown size and bound between important events (Schmidhuber, Wierstra and Gomez, 2005). LSTMs displayed a strong record for addressing problems in speech recognition (Graves, et al. 2013), rhythm learning (Gers, Schraudolph, and Schmidhuber, 2002) etc. However, a disadvantage of using LSTMs (and RNNs in general) is that they are slower with respect to feed-forward networks; such as convolutional neural networks (ConvNets). ConvNets are feed-forward

# Modelling prosodic structure using Artificial Neural Networks

artificial neural networks, which learn patterns in restricted regions a.k.a., receptive fields that partially overlap. These architectures were inspired by biological processes of visual perception. These architectures were successfully employed to address tasks in image recognition (Ciresan, Meier, and Schmidhuber, 2012) and sentence processing.

## Experiments

For the purposes of this study, 1966 statement productions were produced by 25 female speakers and 2860 Wh-questions were produced by 20 female speakers of Cypriot Greek. Sound files were segmented and labelled and the corresponding F0 contours were extracted every 12.5 ms. The data was then padded with zeros for a total length of 1024 and provided as an input to our models. Statements consisted of the same utterance melody "ipes CVCV pali" *you said CVCV again*, where C stands for a consonant and V for a vowel whereas Wh questions varied both in length and in the initial Wh-word. For the purposes of this study a twofold experiment has been employed, namely a classification task with an LSTM neural network and a classification task with a ConvNet.

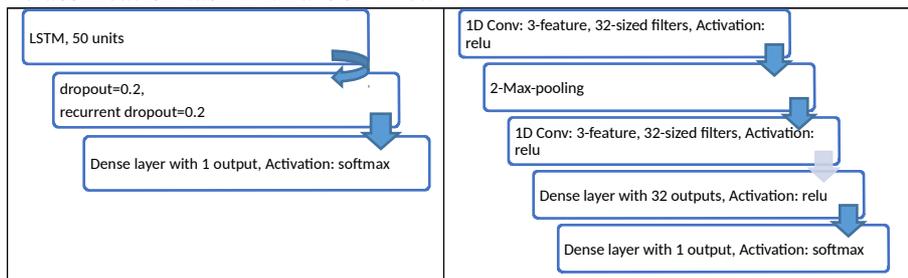

Figure 2. LSTM Neural Network Architecture (left panel) and ConvNet Architecture (right panel).

For each run models were trained for 18 epochs with backpropagation. After each epoch, a model is obtained. Out those 18 models, the model yielding the smallest loss on the training data is kept, and used for evaluation against the test set. The LSTM resulted in 82% classification accuracy, while the ConvNets resulted in over 95% accuracy. For the ConvNet, we performed 10-fold cross validation (4343 training examples and 434 test examples). The distribution of the classification accuracy has median 95%. (Min. 90%, 1st Qu. 94%, Median 95%, 3rd Qu. 96% and Max. 97%).

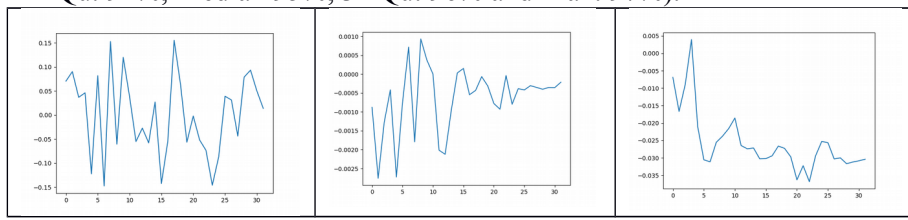

Figure 3. Examples of learned convolutional filters for the first layer.

## Discussion

We have described two artificial neural network architectures: an LSTM and a convolutional network. The ConvNet outperformed the LSTM in accuracy and speed. Specifically, the ConvNet achieves high performance on question and statement classification without requiring external features, such as manually tagged pitch accents. The network can identify by itself the properties of the tonal contour that are most significant for the classification. Despite the high accuracy learned features are not easily interpreted from a linguistic point of view (see Figure 3), yet a promising aspect of this architecture is that it requires few parameters (only 6 filters). However, this study was performed on a corpus that exhibits regularities and it remains to be seen whether it generalizes in a more varied corpus.